\documentclass[10pt,twocolumn,letterpaper]{article}

\usepackage[pagenumbers]{cvpr} 

\usepackage{graphicx}
\usepackage{amsmath}
\usepackage{amssymb}
\usepackage{booktabs}

\usepackage[pagebackref,breaklinks,colorlinks]{hyperref}

\usepackage[capitalize]{cleveref}
\crefname{section}{Sec.}{Secs.}
\Crefname{section}{Section}{Sections}
\Crefname{table}{Table}{Tables}
\crefname{table}{Tab.}{Tabs.}

\def\cvprPaperID{6142} 
\def\confName{CVPR}
\def\confYear{2022}

\begin{document}

\title{QK Iteration: A Self-Supervised Representation Learning Algorithm for Image Similarity}

\author{David Wu\\
{\tt\small david9dragon9@gmail.com}
\and
Yunnan Wu}
\maketitle
\begin{figure*}
  \begin{center}
  \includegraphics[width=\textwidth] {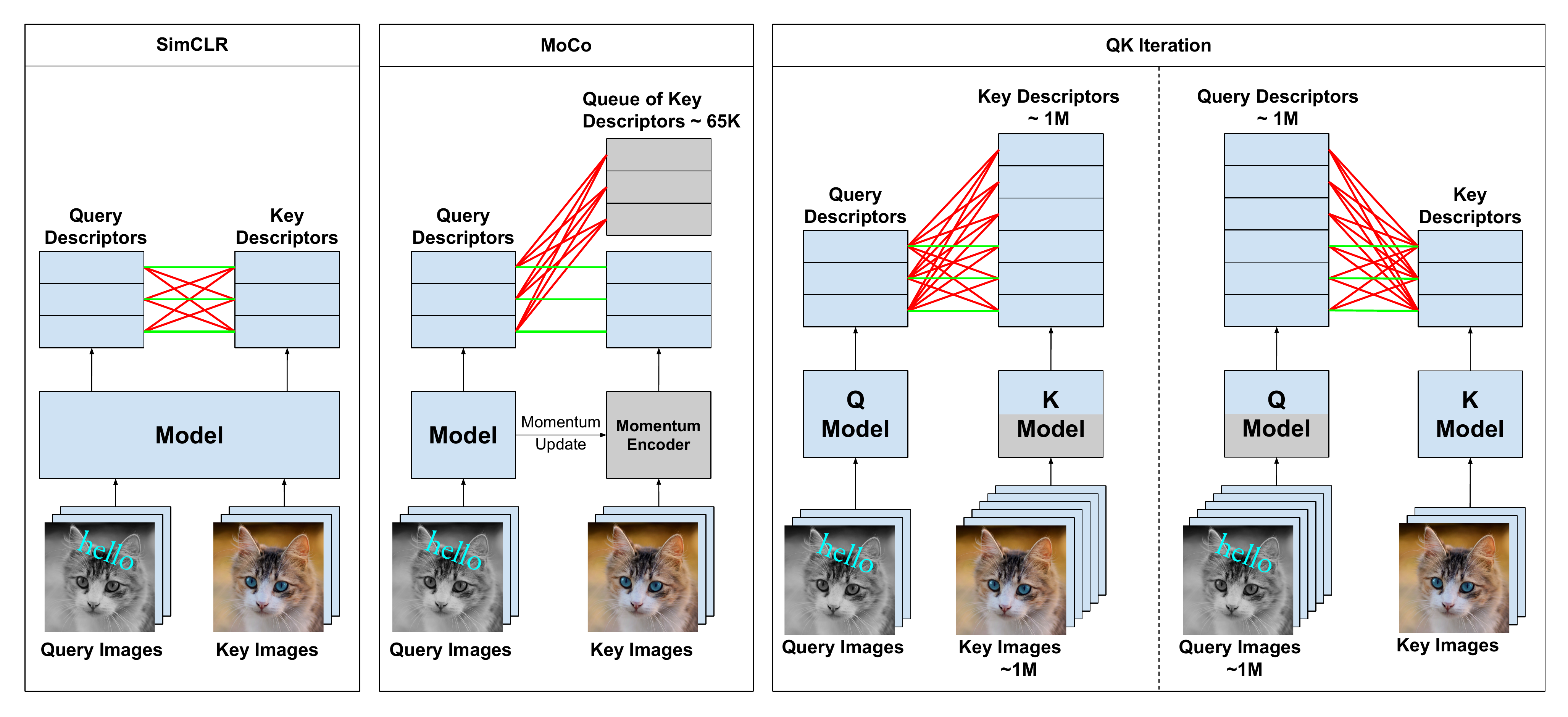}
  \end{center}
     \caption{Illustration of the proposed self-supervised learning algorithm, compared with SimCLR \cite{Chen2020SimCLR} and MoCo \cite{He2019MoCo}. Best viewed in color. Green lines indicate positive pairs, and red lines indicate negative pairs.}
  \label{fig:summary}
  \end{figure*}

\begin{abstract}
  Self-supervised representation learning is a fundamental problem in computer vision with many useful applications (e.g., image search, instance level recognition, copy detection). In this paper we present a new contrastive self-supervised representation learning algorithm in the context of Copy Detection in the 2021 Image Similarity Challenge hosted by Facebook AI Research. Previous work in contrastive self-supervised learning has identified the importance of being able to optimize representations while ``pushing'' against a large number of negative examples. Representative previous solutions either use large batches enabled by modern distributed training systems or maintain queues or memory banks holding recently evaluated representations while relaxing some consistency properties. We approach this problem from a new angle: We directly learn a query model and a key model jointly and push representations against a very large number (e.g., 1 million)  of negative representations in each SGD step. We achieve this by freezing the backbone on one side and by alternating between a Q-optimization step and a K-optimization step. During the competition timeframe, our algorithms achieved a micro-AP score of 0.3401 on the Phase 1 leaderboard, significantly improving over the baseline $\mu$AP of 0.1556. On the final Phase 2 leaderboard, our model scored 0.1919, while the baseline scored 0.0526. Continued training yielded further improvement. We conducted an empirical study to compare the proposed approach with a SimCLR style strategy where the negative examples are taken from the batch only. We found that our method ($\mu$AP of 0.3403) significantly outperforms this SimCLR-style baseline ($\mu$AP of 0.2001).
\end{abstract}

\section{Introduction}
\label{sec:intro}

The 2021 Image Similarity Challenge (ISC)\cite{douze20212021}, hosted by Facebook AI Research (FAIR), was one of the largest competitions on the problem of Copy Detection. Copy Detection is the problem of detecting whether a current query image is an edited version of any image in a database of key images, and which one, if it exists. It is a critically important problem in the real world, with significant impacts. Commercially, solving the problem of Copy Detection could help companies prevent the spread of manipulated media and prevent misinformation, both opportunities to increase users' trust, safety, and general wellbeing. ISC2021 offered a large dataset of reference images, and a smaller dataset of query images, some of which were edited copies of images in the reference database. There were two tracks to the challlenge: 

\begin{itemize}
  \item The Matching Track, where the goal was only to find the matching images, with no restrictions on methods.
  \item The Descriptor Track, where the goal was to extract 256d descriptors for the query and reference images so that a similarity search based on L2 distance would find the matching pairs.
\end{itemize}

We participated in the descriptor track. In this paper we will be focusing on learning similarity-preserving image descriptors for Copy Detection. The gist of this problem, however, is self-supervised learning of image representations. As a result, our proposed methods are self-supervised learning algorithms. In recent years, significant progress has been made in the field of self-supervised representation learning. These methods try to find representations that are invariant to data augmentation. The original motivation for this line of research was to develop pretrained models using self-supervised learning that can be fine-tuned on application-specific data in order to obtain better performance given this prior knowledge on various supervised learning tasks, such as image classification. These algorithms are closely related to the Copy Detection problem.



Our proposed algorithm is a new way of learning descriptors that preserve image similarity in a self-supervised manner. We believe it is a useful contribution to the field of self-supervised representation learning. 

From another perspective, one will notice that the problem of learning descriptors that preserve similarity has a long history, with applications ranging from Instance Level Recognition (e.g. Google Landmarks Challenge\cite{GoogleLandmarksRetrievalChallenge}) to Copy Detection (e.g. Facebook Image Similarity Challenge\cite{douze20212021}). It is a core problem to computer vision, as it forms the basis for Image Retrieval. Image Retrieval is the problem of retrieving an image that satisfies a certain property or has a certain relationship to a current image, out of a database of images. Problems like Copy Detection and Instance Level Recognition are specific cases of the Image Retrieval problem, for example recognizing an edited image or an image of the same object/instance. The focus of this paper will be on Copy Detection, since that is the focus of the 2021 Image Similarity Challenge (ISC2021) hosted by Facebook AI Research (FAIR).

For this task, there are two types of images: query and key images. Key images are the original, non augmented images, whereas query images are the augmented/edited versions of the key images. The task at hand is to generate descriptors for the query and key images, such that those for positive pairs, where one is an edited version of the other, are closer in L2 distance, whereas negative pairs, or non matching pairs of images, are farther away. Our QK Iteration method is a new self-supervised learning algorithm for learning similarity-preserving image descriptors. Figure~\ref{fig:summary} illustrates the proposed self-supervised learning algorithms, in comparison with well known self-supervised learning algorithms SimCLR \cite{Chen2020SimCLR} and MoCo \cite{He2019MoCo}.

The key features for our solution are:
\begin{itemize}
  \item We maintain distinct Q and K models that allow us to better model the distinct distributions of the query and key images \emph{separately}, allowing our model to learn descriptors that are better suited for query and key images specifically. This is reflected in Figure~\ref{fig:summary} by the fact that we explicitly optimize two models.
  \item Previous work \cite{Chen2020SimCLR, He2019MoCo} has shown the importance of pushing against a large and diverse set of negative examples for contrastive self-supervised learning. Unlike previous well-known algorithms for learning descriptors, our QK algorithm is able to push against descriptors for around 1 million images in each SGD step, creating a wider range of negative examples and improving the quality of learned descriptors. This is achieved by freezing the backbone on one side (illustrated by the gray bottom half of the models in Figure~\ref{fig:summary}).
  In contrast, in SimCLR, the model pushes against one batch, and in MoCo, the model pushes against a queue of previously evaluated descriptors, with a loosened consistency guarantee.
  \item We are able to simultaneously train parts of both the query and key models in order to allow coordinated movements and freedom to adjust both sides' descriptors. This is illustrated by the fact that we have blue (indicating trainable) parts in both the Q and K models in Figure~\ref{fig:summary}.
  \item We alternate between training the query model's backbone and the key model's backbone in order to save computational cost, while still providing opportunities for the model to move both the query and the key descriptors \emph{freely}, allowing for quality descriptors on both the query and key sides.
  \item We used a loss function that consists of simple binary cross entropy with hard negative mining that better aligns with the competition metric, which made scores for different query-key pairs more comparable. By using hard negative mining, we prevented negative loss terms from overpowering and adjusted the model using appropriate signals from both the positive and negative pairs.
\end{itemize}

\section{Related Work}
\subsection{Contrastive Self-supervised Learning}
There are many well known approaches for the problem of learning similarity preserving image descriptors, many in the field of self-supervised learning, including SimCLR \cite{Chen2020SimCLR}  and Momentum Contrast \cite{He2019MoCo}.

{\bf SimCLR:}
SimCLR\cite{Chen2020SimCLR} trains one model to produce these descriptors by taking a batch of images and their augmented counterparts, and making the descriptors for matching images to be closer, and those of non matching images to be farther away, using a contrastive loss. The main issue with SimCLR is that it only optimizes each batch of descriptors to match that specific batch, only ``pushing'' against that batch, instead of finding descriptors that optimize for the ``close to positive pairs and far from negative pairs'' property for the entire database of images, so it might take longer to find descriptors that fit well for the entire dataset, and will require very large batch sizes, which are not always available.

{\bf Momentum Contrast(MoCo):}
MoCo addresses this problem of not pushing against a wide enough variety of negative examples by introducing the concept of momentum updates. Instead of only pushing descriptors against other descriptors in the same batch, MoCo pushes them against a maintained queue of previously evaluated descriptors. However, these previously evaluated descriptors will be inconsistent with the current model. MoCo tries to solve this problem by adopting a slowly evolving key model that is an exponentially weighted average of the query model's weights, so that the descriptors in the queue are consistent with each other. However, MoCo does not fully solve the problem of not pushing against enough diverse negatives, instead choosing to relax consistency to be able to push against more descriptors. 

In addition, both of these algorithms have one problem that causes them a disadvantage in the problem of Copy Detection. While they both only learn one model for all images, we actually need to be able to differentiate between query and key images, and treat them differently. For example, one might speculate that focus would need to be placed on salient parts of query images, whereas for key images, one might need to deal with background differently. We choose to train two separate models for query and key images.

\subsection{Copy Detection Baselines}
{\bf GIST:}
The GIST\cite{oliva2001holistic} descriptor is a fixed, non machine-learning based way to extract descriptors from an image. It applies Gabor Filters over the image to build gradient histograms, capturing the overall idea of the image. It is not trainable, and outputs descriptors of length 960. However, as the descriptor track limits the length of the image descriptors to be less than 256, a PCA without whitening is applied to reduce the dimension to 256. GIST is used as the baseline in the descriptor track, achieving a $\mu$AP of 0.1556 on the Phase 1 data.

{\bf MultiGrain:}
The MultiGrain\cite{berman2019multigrain} model is essentially a classical ResNet50, with an extra GeM pooling head that outputs descriptors, along with the typical classification head. The model is then trained to produce similar descriptors for different augmented versions of an image, and produces descriptors of dimension 2048. After applying a PCA with whitening, the length is reduced to 1500, which scores 0.1540 $\mu$AP on the 25000 subset, not much better than GIST. Note that the dimension would need to be further reduced in order to qualify for the official descriptor track.



\section{Proposed Algorithm}

\subsection{Model}
\begin{figure}[t]
  \begin{center}
  \includegraphics[width=0.9\linewidth]{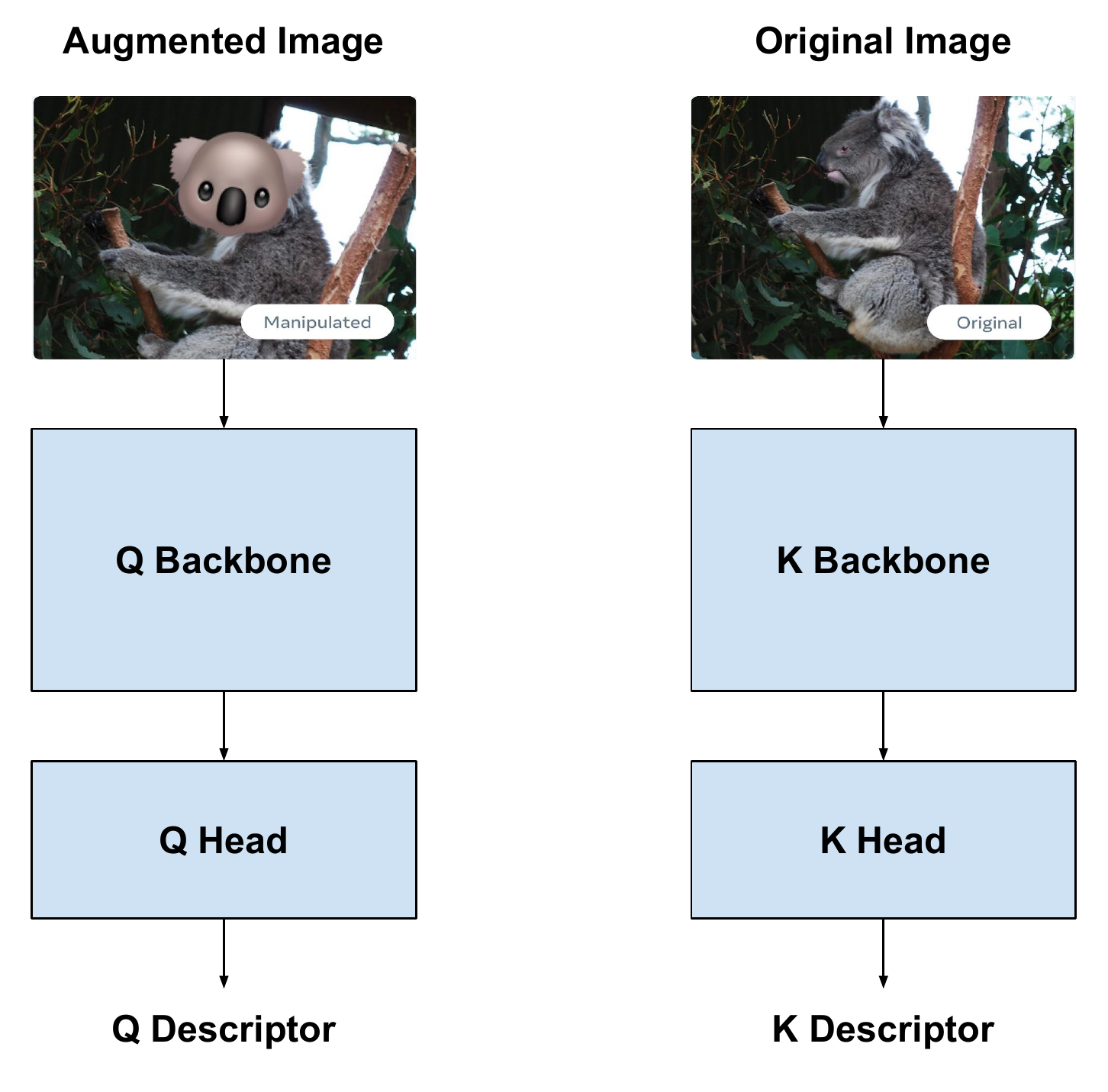}
  \end{center}
     \caption{The model setup. We maintain distinct models for query images and key images, respectively. Example images are taken from the competition website \cite{douze20212021}.}
  \label{fig:model_setup}
  \end{figure}

For the copy detection problem, the query images and the key images are not symmetric and may have different distributions. For example, the query images commonly involve editing and composition operations whereas the key images tend to have less composition. In real world applications, the only requirement is to be able to compute query image representations and compare them against pre-computed key image representations. There is no need to require the query and key representations to be from the same model. Considering the asymmetry, to better model the different distributions of the query and key sides, we decided to train two models, one query model and one key model. This setup is illustrated in Figure~\ref{fig:model_setup}. Within each model, we have a CNN backbone followed by a head. The backbone can just be a common pre-trained CNN backbone, such as an EfficientNet or a ResNet. The head can just be a standard feed-forward network (FFN).



In an ideal world without computational constraints, we would fully train both model Q and model K simultaneously with a contrastive loss, which pushes the representations for positive pairs of (query, key) images close together and those for negative pairs far away. At each step, we would evaluate the key model on the entire database of key images, to obtain a diverse set of descriptors to push against. Then, we would take a batch of query images, and evaluate our query model on them to obtain a set of query descriptors. We would calculate our loss, by pushing the batch of query descriptors against the key descriptors evaluated on the entire dataset, where a pair of query and key images would be a positive pair if the query is an edited version of the key, and negative otherwise. We would then calculate gradients through back propagation and update both the query and key models fully.

However, this solution is infeasible in practice, as we would need to bulk evaluate and update the key model on the entire database of key images at each step, bringing about an unmanageable computational burden, and would be impossible for problems with large datasets, like ISC2021. 

We approach this problem by trying to make simplifications to the above ideal solution to arrive at computationally feasible solutions.


\subsubsection{QK Co-learn Model}

\begin{figure}[t]
  \begin{center}
  \includegraphics[width=0.8\linewidth]{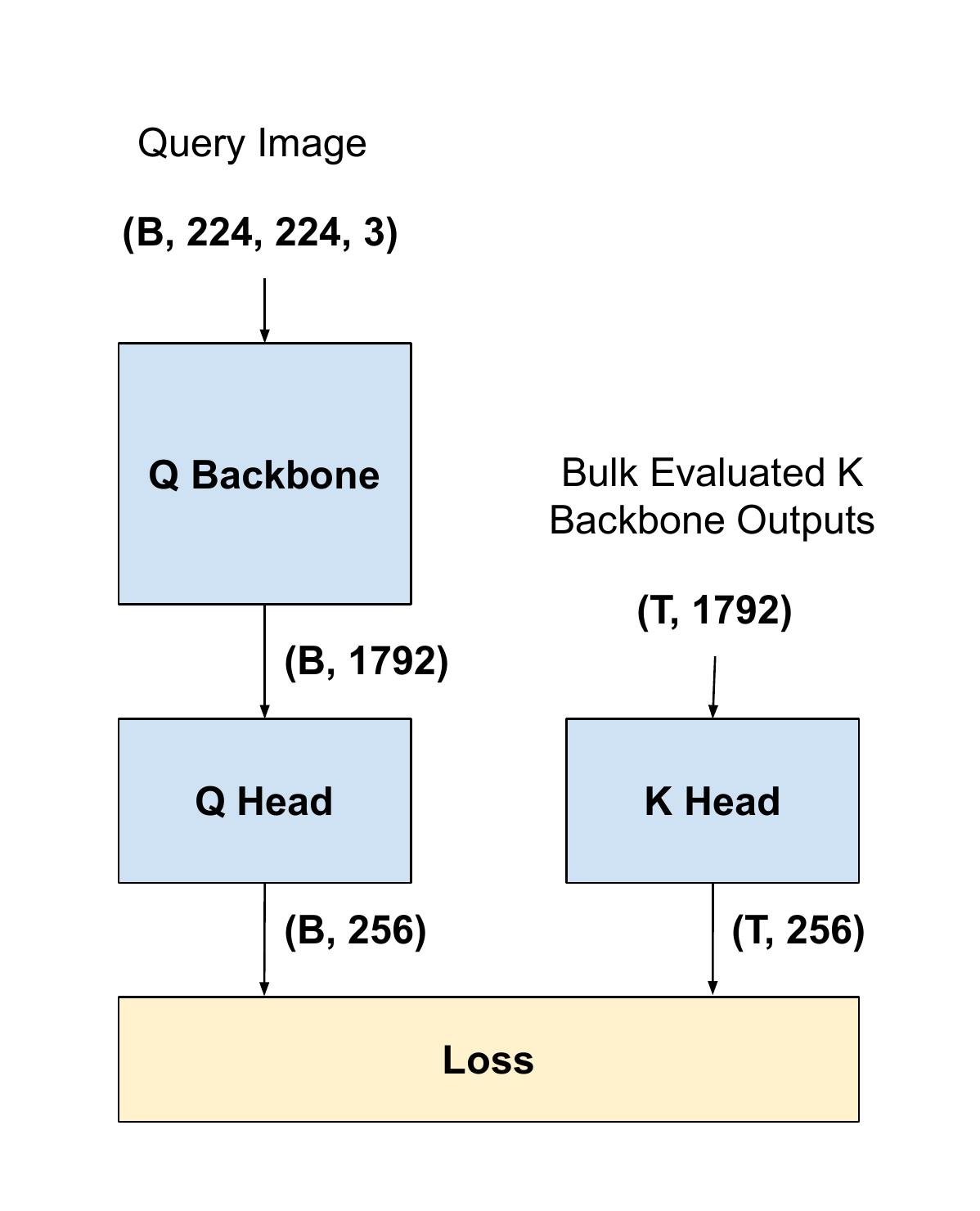}
  \end{center}
     \caption{Our QK Co-Learn Model Setup. B is the batch size, and T is the number of images in the data chunk that we are currently training with.}
  \label{fig:qk_colearn_model}
  \end{figure}

We solve this computational problem by not updating the \emph{full} query and key models after each step, instead freezing the K model's backbone. We then proceed to bulk evaluate the key model's backbone, and obtain intermediate descriptors for the database of key images. In the ISC competition, we compute 1 million length-1792 descriptors for the entire training dataset. This just needs to be done once before the training. In case the training data is much bigger, we can break the dataset down into chunks and train with each chunk individually. Then, as shown in Figure~\ref{fig:qk_colearn_model}, at each step during forward propagation, we would evaluate our key model head on the previously saved intermediate descriptors to obtain final descriptors for the entire database of key images. We evaluate our query model on a batch of query images, as usual, and push the query descriptors against the database of key descriptors. During back propagation, the Q backbone, Q Head, and K Head are updated.

Freezing the K backbone allows us to push against all 1000000 images in the key database, without incurring too big of a computational cost or memory cost. This allows us to push against a more complete set of negative examples and increases the quality of learned descriptors.

However, freezing the K model backbone restricts the freedom of the K model to adjust its descriptors for the database of key images to positions that fit well for the query descriptors. We would like to be able to dynamically adjust the values of both query \emph{and} key descriptors to find good positions for them. In fact, the intermediate descriptors for the key images come typically from a pretrained CNN backbone, which might not be perfectly suited for our task. We would like to be able to train the entire K model.

\subsubsection{QK Iteration}
To solve this problem, we propose the idea of \emph{QK Iteration}, or alternating between training the Q model and freezing the K model's backbone, and training the K model while freezing the Q model's backbone. After training the Q model for a reasonable amount of time using the training process described above, we would switch to training the K model fully.
\begin{enumerate}
  \item First, we would bulk evaluate the trained Q model's backbone to obtain intermediate descriptors for 1 million augmented images, just like we did for the K model.
  \item Then, we would train our K model fully by evaluating our trainable Q head on the intermediate descriptors and push them against evaluated K descriptors for each batch of key images. We would then use Gradient Descent to update the Q model's head and the entire K model.
\end{enumerate}

\begin{figure}[t]
  \begin{center}
  \includegraphics[width=0.95\linewidth]{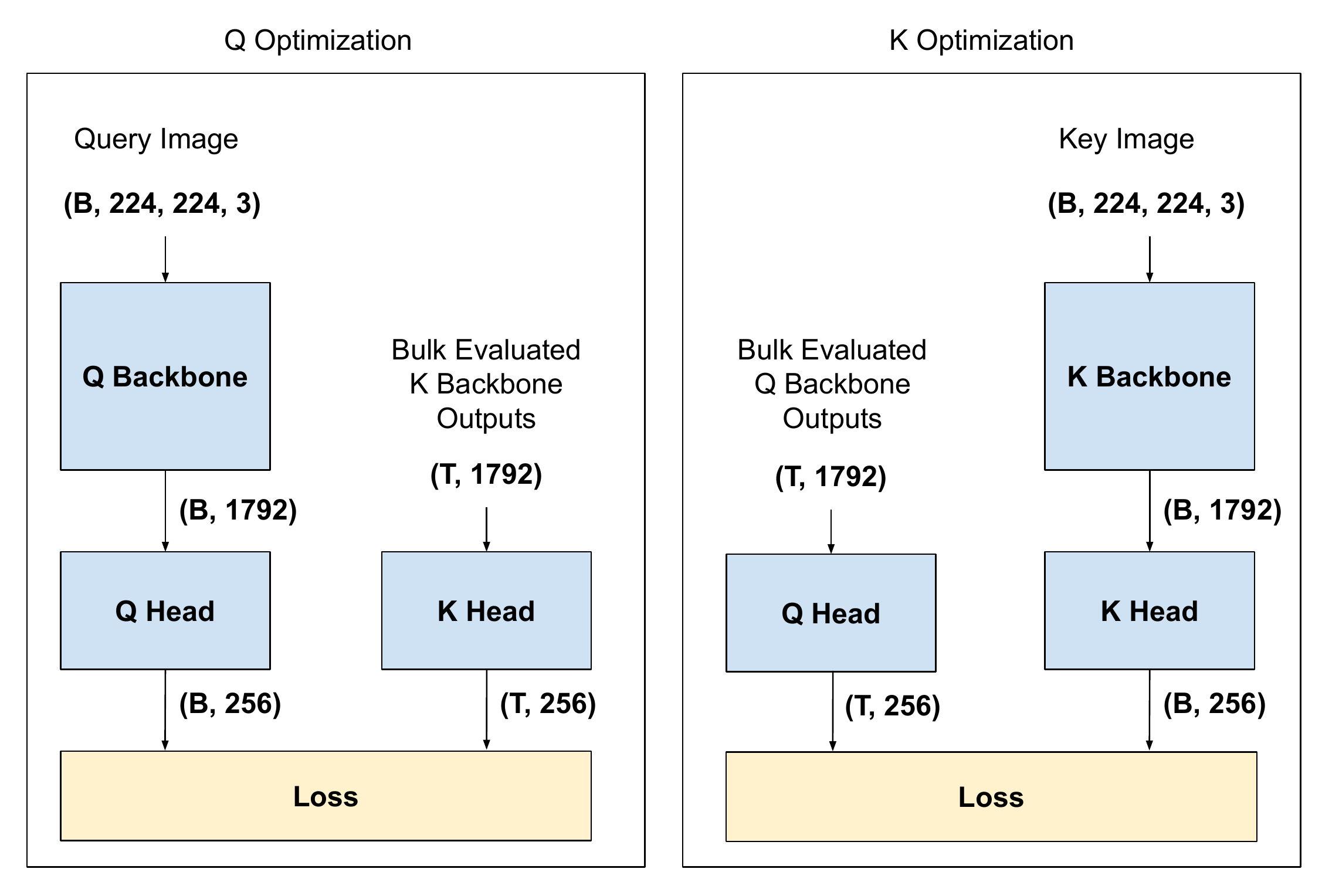}
  \end{center}
     \caption{QK Iteration, illustrated}
  \label{fig:qk_iteration}
  \end{figure}

We would perform this iteration repeatedly, and alternate between bulk evaluating our key model's backbone to focus on our Q model, and bulk evaluating our Q model's backbone to train our K model, as shown in Figure~\ref{fig:qk_iteration}.

This allows us to both train our Q and K models' backbones, still without significantly increasing the computational cost, or giving up the ability to push against a wide range of negative examples, which maintains the quality of our descriptors.

One might see the resemblance to the classical K-Means clustering algorithm, which alternates between two steps: adjusting positions of the cluster centroids, then adjusting assignments of the given datapoints. Similar to the K-Means algorithm, our algorithm alternates between two steps: fully training the Q model and fully training the K model.

The intuition behind this approach is that instead of trying to decrease the computation by decreasing the amount of negative examples or approximating key descriptors, we break the training process down into two parts and alternate between the two processes, allowing the model to work on the two pieces separately, which saves computation without constraining the solution space too much.

Lastly, note that we could also perform QK iteration by completely freezing one side and optimizing the other side, which could further save some computations. However we favor partial freezing over full freezing. The rationale behind only freezing the CNN backbone instead of completely freezing the other model is that we would like to give the model some freedom to adjust key descriptors even while focusing on query descriptors, as it might be hard for the model to find good values for the query descriptors when the key descriptors are frozen at suboptimal values. In addition, movements of the query and key descriptors can be better coordinated. Our intuition is that the added freedom to adjust the key descriptors is worth the additional computational cost of evaluating the head at each step.

\subsection{Loss}
Previous self-supervised learning algorithms, such as SimCLR and MoCo, used the 
InfoNCE\cite{oord2019representation} loss, which is essentially a multi-class log loss that tries to classify a query ${\bf q}$ as the matching key class ${\bf k}_{+}$ instead of the negative key classes. 
However, we felt that InfoNCE was \underline{not} the best loss choice for us, as it does not align well with our metric, $\mu$AP.  Our metric requires distances for all query and key pairs to be comparable, but InfoNCE does not guarantee us that property, as distances are compared only within each single query. 






Instead, we turn the problem into one of classifying positive and negative (query, key) pairs, to better align with $\mu$AP. In our loss, we use a simple binary cross entropy over positive and negative pairs' score predictions, where the score is $e^{\frac{-d^2}{\tau}}$ of the distance $d$. This pushes distances for positive pairs downwards, and those for negative pairs upward. In equations, the loss $L$ is

\begin{align}
P_{ij} &\triangleq e^{\frac{-\left\| {\bf q}_i - {\bf k}_j \right\|^2}{\tau}} \\
L_{pos} &\triangleq \frac{\sum_{i,j\in positives}-log(P_{ij})}{B} \\
L_{neg} &\triangleq \frac{\sum_{i,j\in top\ B*M\ negatives}-log(1-P_{ij})}{B*M} \\
L &\triangleq w_{pos}*L_{pos} + w_{neg}*L_{neg}
\end{align}
where ${\bf q}_i$ and ${\bf k}_j$ are the query and key descriptors, respectively. If we are currently training the Q model, the $i$ index would index the samples in the batch, and the $j$ index would index the database images, and vice versa if we are currently training the K model. In addition, $B$ is the batch size, $M$ is the parameter controlling the number of hard negatives per positive, and $\tau$ is the temperature. $w_{pos}$ and $w_{neg}$ are weights for the positive and negative parts of the loss. In all our experiments, $\tau = 0.07$, $M = 10$, $w_{pos}=1$, and $w_{neg}=3$.

Lastly, because we had so many negative pairs, we need to apply some treatment to avoid the negative pairs overpowering the positive pairs. We chose to use hard negative mining\cite{DBLP:journals/corr/ShrivastavaGG16}, by taking the top $B*M$ negative pairs with smallest distance to calculate the negative part of the loss. As a future exploration, we could also consider using Focal Loss\cite{lin2017focal}.

\section{Technical Details}

\begin{figure}[t]
  \begin{center}
  \includegraphics[width=0.8\linewidth]{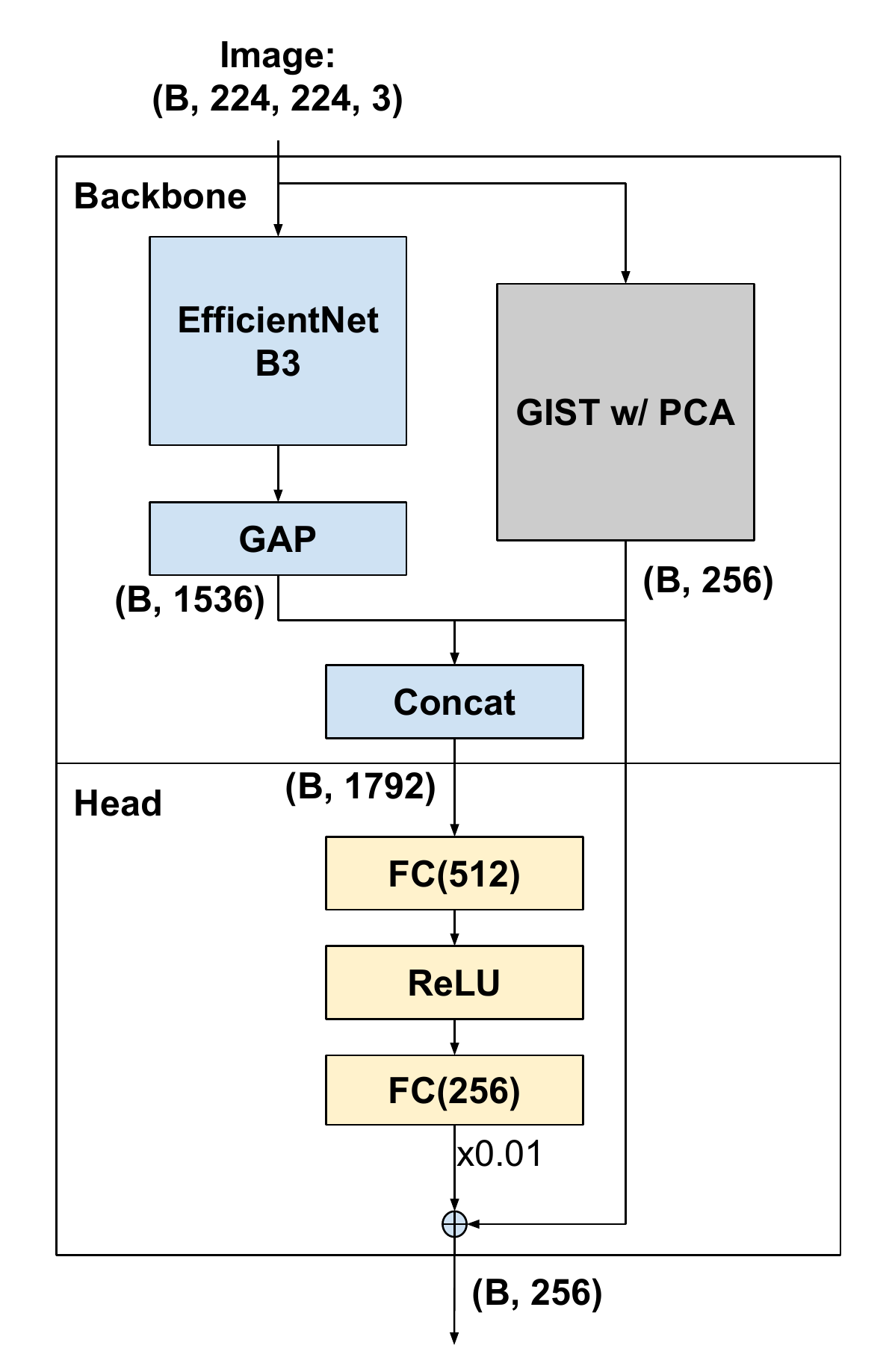}
  \end{center}
     \caption{Our detailed model architecture}
  \label{fig:detailed_arch}
  \end{figure}

\subsection{Detailed Model Architecture}
Figure~\ref{fig:detailed_arch} illustrates the detailed model architecture. The same backbone and head architecture is adopted for both Q and K models. For the CNN backbone, we used EfficientNet B3, pretrained on ImageNet. 

In order to give our model a good starting point, we incorporated the GIST baseline into our model, as shown in Figure~\ref{fig:detailed_arch}. First, we modified the final output by multiplying it by 0.01 and adding on the GIST descriptors. Note that we followed the baseline recommendation in \cite{douze20212021} and reduced the original 960d descriptors to 256d using PCA without whitening, in order to make them compatible with our output descriptors. In addition, we concatenated it with the intermediate descriptors from the backbone before feeding it in to the final head. Hence, our model has a good starting point and does not spend too much time trying to get to a good initialization. In addition, concatenating the GIST descriptor onto the intermediate descriptor also prevents our model from being too constrained and dominated by GIST, as it can dampen the effects of GIST by cancelling it out if needed. Lastly, as the GIST descriptor is relatively inexpensive to calculate, this does not add too big of a computational cost.

{\bf Batch Normalization:} 
In our experiments, we observed that the use of Batch Normalization led to good training metrics but poor evaluation metrics. We suspect this is due to information leakage, as our loss function had intrabatch communication by taking the top k negative examples in each batch. This is consistent with the findings reported in \cite{Chen2020SimCLR,He2019MoCo}. To solve this problem, we simply set the trainable attribute in the Batch Normalization layers to $\mathtt{False}$. We defer exploration of more complex treatments to future work.

{\bf When to switch?}
A simple strategy for determining when to switch from training Q to K, or K to Q would be to wait for performance to saturate. However, this may not be the optimal strategy. Alternatively, switching more often might allow for more flexibility in navigating the solution space. This is an interesting topic for future investigation.

\subsection{Dataset and Augmentation}

We trained the model using the provided training dataset containing 1 million images. To generate the query images, the key images were augmented using randomly chosen compositions of augmentations implemented using the Facebook AugLy augmentation library\cite{bitton2021augly}. Augmentations were only applied to obtain the query images. For the key images, we used the provided set of images. The exact transformations that we used can be found in our code, which we will release on GitHub. Each of the 1000000 training key images was augmented to generate its corresponding query image, so that the (query, key) pair generated is a positive pair. Note that this is an example of self-supervised learning, as the model is generating labels by itself.

\subsection{Mixed Precision Computing}
One of the problems that we encountered when training our model was that fitting the preevaluated descriptors for the entire database of images took up a lot of memory, and didn't allow us to fully unfreeze the EfficientNet backbone. To solve this problem, we stored the preevaluated database descriptors as $\mathtt{float16}$s, or half precision floating points to save memory, and used Mixed Precision capabilities provided by TensorFlow to feed it into the head.

\section{Experiments}
\subsection{Setup}
For our experiments, we used an ImageNet pretrained EfficientNetB3 as our CNN backbone provided by $\mathtt{tf.keras.applications.EfficientNetB3}$. Images were resized to (224, 224, 3) to feed into the EfficientNet. Our solution was implemented using TensorFlow.

\subsection{Training}
We used a training batch size of 32, and trained our model using an Adam optimizer \cite{Adam} with default settings. For the learning rate, we used a Cosine Decay schedule implemented with $\mathtt{tf.keras.optimizers.schedules.CosineDecay}$ with an initial learning rate of 0.0001, $\alpha=0.5$ decaying to half over 31250 steps (i.e., one pass over the 1 million images). Models were trained on Google Colaboratory, which provides 1 NVIDIA Tesla P100 GPU with 16 GB of GPU RAM or 1 NVIDIA Tesla T4 GPU with 15 GB of GPU RAM. We trained on the 1000000 training images provided by the ISC competition organizers, which contains images from the YFCC100M dataset and the DFDC challenge organized by Facebook. When we had access to the P100 GPU, we directly pushed against 1 million examples. When only the T4 GPU was available, we split the dataset into two halves, each containing 0.5 million examples and corresponding preevaluated descriptors, and proceeded to train on the two halves one by one. 

Due to limited access to computational resources, we had to scale down our experiments and were not able to fully optimize our models or do extensive parameter tuning. We believe that our method has yet to reach its full potential due to the small number of epochs and small scale. In the future, we believe that our algorithm can achieve better performance given more computational resources.

\subsection{Metric}
The competition's primary metric is $\mu$AP. $\mu$AP is computed by sorting all of the matched distances, and then calculating the area under the PR curve by calculating precision and recall at different thresholds. This is different from the macro-AP, which calculates the area under the PR curve by ranking results for each query, then averages over all queries. $\mu$AP requires the distances between different (query, key) pairs to be comparable. We evaluated our models that were trained on the training dataset on a completely separate set of query and reference images provided by the competition. 

\subsection{Competition}
During the competition, as we were continuously experimenting with model configs and resumed training from the best snapshots to save time, our competition model architecture was more complicated. In retrospect, we recommend a cleaner architecture as in Figure~\ref{fig:detailed_arch}; in the next subsection we describe the updated results that we retrain from scratch with Figure~\ref{fig:detailed_arch}.
The architecture of our submitted model included an extra FFN with a residual connection after the head shown in Figure~\ref{fig:detailed_arch}. More specifically, if we denote the resulting $(B,256)$ tensor shown in the figure as ${\bf e}$, then the final descriptor is $0.01*f_1(ReLU(f_2({\bf e}))) + {\bf e}$, where $f_1$ and $f_2$ are standard dense layers with 256 units.

The results for the competition can be found in Table~\ref{table:competition_results}. Here the first column indicates the training phase; for example, K1 means after finishing the first K optimization. The second column shows the $\mu$AP over the 25000 examples where the ground truth was given. The third column shows the $\mu$AP over the 50000 examples; this corresponds to the Phase 1 leaderboard results. 
For Q1 and K1, we were using the LARS optimizer \cite{DBLP:journals/corr/abs-1708-03888}. Later, we found that the Adam optimizer worked better than the LARS optimizer. Thus the subsequent optimization steps were using the Adam optimizer.

\begin{table}[ht]
  \caption{Competition Results} 
  \centering 
  \begin{tabular}{c  c c} 
  \hline\hline 
  Phase & $\mu$AP: 25K examples & $\mu$AP: 50K examples \\ [0.5ex] 
  \hline 
  Q1, K1 & 0.1908 & 0.1953 \\
  Q2 & 0.2813 & 0.2877  \\
  K2 & 0.3049 & 0.3139  \\
  Q3 & 0.3328 & 0.3401 \\[1ex] 
  \hline 
  \end{tabular}
  \label{table:competition_results} 
\end{table}

On the final Phase 2 leaderboard, our model scored 0.1919, whereas the baseline method scored 0.0526.

\subsection{Updated Results}
After the competition, we trained a new model from scratch using a cleaner model architecture (Figure~\ref{fig:detailed_arch}) and with the Adam optimizer throughout. The updated results can be found in Table~\ref{table:ideal_results}. We report $\mu$AP on the 25000 provided ground truth examples in the evaluation dataset. Each training step runs for about 2 epochs.

\begin{table}[ht]
  \caption{Updated Model Results} 
  \centering 
  \begin{tabular}{c  c} 
  \hline\hline 
  Phase & $\mu$AP: 25K examples \\ [0.5ex] 
  \hline 
  Q1 & 0.2430\\
  K1 & 0.2839\\
  Q2 & 0.3191\\
  K2 & 0.3295\\
  Q3 & 0.3403\\[1ex] 
  \hline 
  \end{tabular}
  \label{table:ideal_results} 
\end{table}

\subsection{Comparison with SimCLR}
We trained an adapted version of SimCLR as a baseline in order to compare performance with our proposed approach. It was trained using two separate query and key models with our loss function and model architecture (with the same GIST residual method), in order to fairly compare the two strategies of pushing against negative examples while keeping all other variables as identical as possible. In order to increase the batch size, as having a large batch size is a crucial part of the SimCLR algorithm, we trained it on TPUv2-8 provided by Google Colab. The batch size was increased to 192, 24 per TPU core. The learning rate was set to 0.001 with decay of $\alpha = 0.5$ over 31250 steps, with the Adam optimizer. We do an all-reduce step to exchange the key side representations of shape (24, 256) among all the cores. After that, each core computes the loss by taking its corresponding (24, 256) shape query representations and pushing against the (192, 256) shape global key representations across the entire batch. Then, the same loss function is used. The SimCLR baseline reached a $\mu$AP of 0.2001 (around 2 epochs) before declining, which is significantly worse than our QK Iteration method ($\mu$AP 0.3403), showing the effectiveness of our strategy of pushing against a large set of negative examples via QK Iteration. This is aligned with prior findings that it is important to push against a large set of negative examples for contrastive self-supervised learning.

\section{Conclusion and Future Work}
We propose a new way of training descriptor models for problems with two sides of images (e.g. Query and Key) called \emph{QK Iteration}, that allows us to effectively push against a wide variety of negative examples (e.g. 1 million) when optimizing descriptors. We train our models by alternating between focusing on training model Q and focusing on training model K, and freezing the backbone of the other model. We optimize a binary cross entropy loss function with hard negative mining, by pushing against the 1 million descriptors generated by the other model.

In the future, it would be interesting to apply the proposed algorithm to the same tasks that SimCLR and MoCo have focused on and compare the performances.


{\small
\bibliographystyle{ieee_fullname}
\bibliography{egbib}
}

\end{document}